\documentclass{article}

\usepackage[final]{corl_2019} 
\usepackage{multirow}

\usepackage{amsmath}
\usepackage{amsfonts}
\usepackage{subfig}
\usepackage{graphicx}

\usepackage{algorithm}
\usepackage{algorithmic}

\usepackage{color}
\usepackage{wrapfig}

\DeclareMathOperator*{\argmin}{\arg\!\min}
\DeclareMathOperator*{\argmax}{\arg\!\max}

\title{Unifying Variational Inference and PAC-Bayes for Supervised Learning that Scales}

\author{
  Sanjay Thakur\\
  Mila\\
  \texttt{thakursa@mila.quebec} \\
  \And
  Herke Van Hoof \\
  UVA \\
  \texttt{h.c.vanhoof@uva.nl} \\
  \AND
  Gunshi Gupta \\
  Mila \\
  \texttt{gunshigupta9@gmail.com} \\
  \And
  David Meger \\
  MRL \\
  \texttt{david.meger@gmail.com} \\
}

\begin{document}
\maketitle


\begin{abstract}
Neural Network based controllers hold enormous potential to learn complex, high-dimensional functions. However, they are prone to overfitting and unwarranted extrapolations. PAC Bayes is a generalized framework which is more resistant to overfitting and that yields performance bounds that hold with arbitrarily high probability even on the unjustified extrapolations. However, optimizing to learn such a function and a bound is intractable for complex tasks. In this work, we propose a method to simultaneously learn such a function and estimate performance bounds that scale organically to high-dimensions, non-linear environments without making any explicit assumptions about the environment. We build our approach on a parallel that we draw between the formulations called \emph{ELBO} and \emph{PAC Bayes} when the risk metric is \emph{negative log likelihood}. Through our experiments on multiple high dimensional \emph{MuJoCo} locomotion tasks, we validate the correctness of our theory, show its ability to generalize better, and investigate the factors that are important for its learning. The code for all the experiments is available at \url{https://bit.ly/2qv0JjA}.
\end{abstract}

\keywords{PAC Bayes, Variational Inference} 


\section{Introduction}\label{sec:introduction}
Supervised learning is one of the most mature techniques in machine learning which learns an input to output mapping in a way that explains the data. Since the rise of neural networks complemented with massive datasets, it has perhaps become the most successful way to have deployable data-driven models~\citep{redmon2018yolov3, van2016wavenet}. The ability to learn from data through neural networks has already seen big successes even in complex applications such as self-driving cars~\cite{pan2018agile, bojarski2016end}. However, methods based on supervised learning are prone to unwarranted extrapolations. The worst case of this unwarranted extrapolation is usually unknown. This is bad for applications where the worst case performance of the models can be catastrophic. The source of this problem is often ascribed to overfitting. Probabilistic models partially mitigate this problem. Most successful and popular supervised learning techniques generate point-estimates. Probabilistic models, on the other hand, have a model-averaging and a regularizing effect that leads to better generalization. But the problem of unwarranted extrapolation still remains at large as can be summarized nicely as \emph{all models are wrong, but some are useful}~\citep{box1987empirical}. 

One solution to this problem is to have data-driven models generate data-driven performance bounds for the unseen data points which is also known as the \emph{generalization risk}, that hold with high-probability, while also mitigating the effects of overfitting as much as possible. Datasets are often limited in many domains like robotics and personalized-treatment. As a result, overfitting has more far-reaching consequences in such problems. For example, a self-driving car which has been trained using a demonstrator who drives safely in the center of the lane might not be able to handle driving at the edge of the road. In such problems, having a measure of generalization risk bounds can be a useful metric that can be used for garnering important information for model deployment. Other examples include, estimating the worst prediction of a cancer predicting system before deploying it in a hospital.

PAC-Bayes~\cite{mcallester1999some} provides a data-driven bound on the generalization performance of a model where the model uses a distribution to make predictions. However, learning probabilistic model that can scale to thousands of data-points, high-dimensional, complex, non-linear tasks is hard. Lately, \emph{variational inference~(VI)} has been used to approximate complex distributions successfully~\citep{blundell2015weight, thakur2019uncertainty, kingma2015variational, kingma2013auto}. VI uses a proxy distribution in place of the true distribution of interest. This true distribution is hard to compute, but the proxy is easy to estimate. But the learned models do not come with any metric for the generalization guarantees.

In this work, we propose a method to develop data-driven models that can scale easily to the high-dimensional and complex non-linear problem, while also being able to estimate the generalization risk. We build our method using the principles of Variational Inference~(VI) on top of the parallels that we draw between a formulation called \emph{Evidence Lower BOund} or \emph{ELBO} and the PAC Bayes formulation with \emph{negative log likelihood~(nll)} as the risk metric. The formulation of \emph{ELBO} makes it invaluable for VI as maximizing \emph{ELBO} is equivalent to minimizing the $\mathrm{KL}$ divergence between the true distribution of interest and its proxy distribution that VI uses. This is a crucial result for many applications where this distribution of interest is used to make predictions. The contribution of our work are
\begin{enumerate}
    \item establishing a relationship between the \emph{ELBO} and PAC Bayes formulation with \emph{nll} as the risk metric,
    \item formulating a method based on \emph{VI} to simultaneously learn data-driven probabilistic models and generalization performance bounds, that can scale organically to high-dimensional problems and hold with arbitrarily high probability.
\end{enumerate}

Through our experiments on complex, high-dimensional \emph{MuJoCo} locomotion tasks, we validate our developed approach, and show its strong generalization potential. We also do a sensitivity analysis of our method concerning its learnability based on certain user-defined settings. Note that our method works without making any explicit assumptions about the environment.

In the next section~(section~\ref{sec:background}), we briefly describe the concepts that we have used in our work. We then cover some related work in section~\ref{sec:related_work} followed by our methodology in section~\ref{sec:methodology}. We show its promise on a few MuJoCo based high-dimensional, complex non-linear locomotion environments in section~\ref{sec:experiments_n_results} and finally conclude with final remarks and future line of work in section~\ref{sec:conclusion_n_future_work}.

\section{Background}\label{sec:background}
In this section, we will first describe our notation, followed by the different pieces that we bring together to build our work.

We define $\mathcal{D} = {(x_i, a_i)}_{i=1}^{N} \in (\mathcal{X} \times \mathcal{A})^{N}$ as the dataset that is available for training where $\mathcal{X}$ and $\mathcal{A}$ correspond to the input-space and output-space respectively. Let $\pi \in \Pi$ be a function in a family of hypothesis defined as $\pi: \mathcal{X} \xrightarrow{} \mathcal{A}$ and $p(\pi)$ defines a distribution over $\pi$. Let $\boldsymbol{w} \in \mathcal{R}^d$ be the learnable or hidden parameters of any learner function distribution. 

\subsection{Generalized Bayes}
One way to generate uncertainty in the prediction in the light of $\mathcal{D}$ and prior beliefs is to estimate the whole posterior distribution by doing bayesian inference. Doing a bayesian inference uses Bayes rule to estimate a posterior distribution of the parameters $\boldsymbol{w}$, i.e. $p(\boldsymbol{w}|\mathcal{D}) = \frac{p(\mathcal{D}|\boldsymbol{w}).p(\boldsymbol{w})}{p(\mathcal{D})}$, where $p({\boldsymbol{w}}|\mathcal{D})$ is the posterior parameter distribution, $p(\mathcal{D}|\boldsymbol{w})$ is the data likelihood, $p(\boldsymbol{w})$ is the prior parameter distribution, and $p(\mathcal{D})$ is the evidence. Usually, there is trade-off between $p(\mathcal{D}\mid \boldsymbol{w})$ and $p(\boldsymbol{w})$. $p(\mathcal{D}\mid \boldsymbol{w})$ aligns the the learned distribution to $\mathcal{D}$ whereas $p(\boldsymbol{w})$ brings a regularizing effect to it. Bayesian inference is different from a lot of supervised learning techniques which either do a \emph{maximum likelihood estimate~(MLE)}, i.e. $\boldsymbol{w}^{MLE} = \argmax_{\boldsymbol{w}} \log p(\mathcal{D}|\boldsymbol{w})$ or a \emph{maximum a posteriori~(MAP)} point estimates $\boldsymbol{w}^{MAP} = \argmax_{\boldsymbol{w}} \log p(\boldsymbol{w}|\mathcal{D})$. Both MLE and MAP are either prone to overfitting or unwarranted extrapolation or both. This problem is easily visible when the seen data does not span the whole space of interest. Hence it is useful to generate predictive distributions instead of mere point estimates. This predictive distribution ideally should have less certainty about its decisions that are far away from $\mathcal{D}$. Predictions at new points are made as $\boldsymbol{x}_*$ as $p(\boldsymbol{a}_*|\boldsymbol{x}_*) = \int p(\boldsymbol{a}_*|\boldsymbol{x}_*,\boldsymbol{w})p(\boldsymbol{w}|\mathcal{D})d\boldsymbol{w}$. However, inferring both $p(\boldsymbol{w}\mid \mathcal{D})$ and $p(\boldsymbol{a}_*|\boldsymbol{x}_*)$ are intractable at times and hard at best. Hence, the usual way to go about them is to approximate them. This is important in a lot of applications where the learnt data-driven function $p(\pi|\mathcal{D})$ comes from the posterior parameter distribution $p(\boldsymbol{w}\mid \mathcal{D})$.

Lately, new techniques based on the notion of \emph{tempered likelihood} has been used~\citep{zhang2006ɛ, zhang2006information, grunwald2012safe, grunwald2016fast}, where the term $p(\mathcal{D}|\boldsymbol{w})$ is replaced with any term that can measure the quality of prediction of learnt function $p(\pi|\mathcal{D})$ on $\mathcal{D}$ or its \emph{empirical risk} that has been trained on $\mathcal{D}$ itself. We denote such \emph{empirical risk} by $l_{\mathcal{D}}(p(\pi \mid \mathcal{D}))$. We define \emph{empirical risk} as $l_{\mathcal{D}}(p(\pi \mid \mathcal{D})) = \frac{1}{N} \sum_{x \sim \mathcal{D}} \mathbb{E}_{\pi \sim p(\pi \mid \mathcal{D})}\left[ \mathcal{L}(\pi(x)) \right]$. Here, $\mathcal{L}$ is a loss that $\pi$ induces on an input $x$. The resulting posterior is called \emph{generalized bayesian posterior} and the process of inferring it is called \emph{generalized bayes}.

\subsection{PAC-Bayes}
PAC-Bayes~\citep{shawe1997pac, mcallester1999some, mcallester1999pac} is a principled framework that is used to generate an upper bound on the \emph{generalization risk}~($l_{\mathcal{X}}(p(\pi \mid \mathcal{D}))$) with an arbitrary probability $1-\delta$ using an \emph{generalized bayesian posterior}. We formalize \emph{generalization risk} as $l_{\mathcal{X}}(p(\pi \mid \mathcal{D})) = \mathbb{E}_{x \sim p(\mathcal{X})} \mathbb{E}_{\pi \sim p(\pi \mid \mathcal{D})}\left[ \mathcal{L}(\pi(x)) \right]$. PAC-Bayes is an extension of \emph{PAC~(Probably Approximately Correct)}~\citep{valiant1984theory} framework to \emph{generalized bayesian posterior}, where the upper bound on $l_{\mathcal{X}}(p(\pi \mid \mathcal{D}))$ is usually the sum of $l_{\mathcal{D}}(p(\pi \mid \mathcal{D}))$, a complexity measure $\mathcal{C}$ between $p(\pi \mid \mathcal{D})$ and $p(\pi)$ and some constants~($\mathcal{K}$) as shown in equation~\ref{eq:PAC_Bayes_definition}.
\begin{equation}\label{eq:PAC_Bayes_definition}
p[l_{\mathcal{X}}(p(\pi \mid \mathcal{D})) \leq l_{\mathcal{D}}(p(\pi \mid \mathcal{D})) + \mathcal{C}(p(\pi \mid \mathcal{D}), p(\pi)) + \mathcal{K}] = 1 - \delta.
\end{equation}
The optimization is done with respect to the distribution over the model parameters to make this bound as tight as possible. The complexity term $\mathcal{C}$ has a regularizing effect, use of $p(\pi\mid \mathcal{D})$ average the results of multiple individual policies and hence tend to perform better and $p(\pi)$ allows to incorporate prior knowledge.

\subsection{Variational Inference}\label{background:VI}
It is often not possible to compute many posterior distributions of interest either due to lack of a closed form solution or due to the high dimensionality of the solution space. In that case we restrict ourselves to a simpler family of distributions which are simpler to evaluate and try to find the best approximation of our target distribution within this family. 
The usual sequence of steps involved are, 
\begin{itemize}
\item Posit a family of approximate densities $\mathcal{Q}$, parameterized by variational parameters $\boldsymbol{\phi}$,
\item Find $q(\boldsymbol{w}\mid \boldsymbol{\phi})$ within $\mathcal{Q}$  that minimizes KL divergence with true posterior density $p(\boldsymbol{w}|\mathcal{D})$, i.e. $q(\boldsymbol{w}\mid \boldsymbol{\phi}^{*})=\argmin_{q(\boldsymbol{w}\mid \boldsymbol{\phi}) \in \mathcal{Q}} \mathrm{KL}\left[q(\boldsymbol{w}\mid \boldsymbol{\phi})||p(\boldsymbol{w}|\mathcal{D})\right]$.
\end{itemize}
It can be shown that minimizing $\mathrm{KL}\left[q(\boldsymbol{w}\mid \boldsymbol{\phi}) || p(\boldsymbol{w}|\mathcal{D})\right]$ is equivalent to decreasing the negative of another formulation called \emph{Evidence Lower Bound~(ELBO)} as shown in equation~\ref{eq:VI_optimization}.
\begin{equation}\label{eq:VI_optimization}
    \begin{split}
        \boldsymbol{\phi^*} &= \argmin_{\boldsymbol{\phi}} \mathrm{KL}\left[q(\boldsymbol{w}\mid \boldsymbol{\phi}) || p(\boldsymbol{w}|\mathcal{D})\right]\\
        &= \argmin_{\phi} \int q(\boldsymbol{w}\mid \boldsymbol{\phi}) \log \frac{q(\boldsymbol{w}\mid \boldsymbol{\phi})}{p(\boldsymbol{w}|\mathcal{D})}\\
        &= \argmin_{\boldsymbol{\phi}} -\mathbb{E}_{q(\boldsymbol{w}\mid \boldsymbol{\phi})}\left[\log p(\mathcal{D}|\boldsymbol{w}) \right] + \mathrm{KL}\left[q(\boldsymbol{w}\mid \boldsymbol{\phi}) || p(\boldsymbol{w})\right].
    \end{split}
    \end{equation}
The obtained formulation in equation~\ref{eq:VI_optimization} is the negative of the \emph{ELBO}. The optimized approximate variational distribution $q(\boldsymbol{w}\mid \boldsymbol{\phi}^{*})$ is then used as proxy for $p(\boldsymbol{w}\mid \mathcal{D})$. The prediction is done as $p(\boldsymbol{a}_*|\boldsymbol{x}_*) = \int p(\boldsymbol{a}_*|\boldsymbol{x}_*,\boldsymbol{w})q(\boldsymbol{w}\mid \boldsymbol{\phi}^*)d\boldsymbol{w}$.



\section{Related Work}\label{sec:related_work}
PAC-Bayes is a generic tool to derive generalization bounds and has been successfully applied in many ML settings~(see \citet{guedj2019primer} for an overview), including statistical learning theory~\citep{maurer2004note, mcallester1999some}, domain adaptation~\citep{germain2013pac, blitzer2008learning, germain2016new}, lifelong learning~\citep{pentina2014pac}, non-iid or heavy tailed data~\citep{alquier2018simpler, ralaivola2009chromatic}, sequential learning~\citep{gerchinovitz2011prediction, li2018quasi}, deep neural networks~\citep{dziugaite2017computing, neyshabur2017exploring}. 
A few pieces of work~\citep{germain2016pac, grunwald2012safe, zhang2006information} have shown that optimizing PAC-Bayes bound with \emph{nll} as risk metric also improves bayesian \emph{marginal likelihood} but they do not suggest a way to learn the distribution over a function that can be used for complex, high-dimensional and non-linear tasks that are common in the many domains. Although Gibbs posterior arises as a natural solution to many PAC-Bayesian bounds, it can be hard to compute such a distribution.

A few works~\cite{alquier2016properties, alquier2017concentration} have used PAC-Bayes to theoretically prove the consistency of variational inference when the loss is the negative log-likelihood. For example, \cite{alquier2017concentration} has proven the consistency of gaussian variational bayes when a parametric model has a log-Lipschitz likelihood and proposed a way to obtain concentration rates for variational bayes to assess the frequentist guarantees for a bayesian estimator. \cite{alquier2016properties} has shown such a theory holds for classification and regression problems. However, computing such quantities can be hard for many problems, such as in robot vision. In our work, we propose and demonstrate a practical technique that scales organically to high-dimensional inputs and the number of data-points.

Probabilistic models have been used at multiple works~\citep{thakur2019uncertainty, higuera2017adapting, higuera2018synthesizing, deisenroth2011pilco, gal2016improving, thakurvery}. However, unlike them we propose a technique that also scalably approximates a PAC-Bayes bound.



\section{Methodology}\label{sec:methodology}
In this section, we first show that using \emph{negative log likelihood~(nll)} as the PAC-Bayes risk metric is algorithmically equivalent to \emph{ELBO} formulation. We define \emph{nll} as $- \log p(\mathcal{D}\mid \boldsymbol{w}) = -\sum_{i=0}^{N}\log p(a_i \mid x_i, \boldsymbol{w})$. Hence, optimizing for \emph{ELBO} optimizes our bound. This allows us to formulate a novel strategy to learning PAC-Bayes generalization bounds using VI. We then describe the training and inference processes followed by a way to extract the generalization bounds.

There have been multiple bounds developed over the years. We choose a bound that allows unbounded loss functions such as \emph{nll} to be used as the risk metric and contains known terms or terms which come from the dataset, so that it can be computed. We build on top of a bound developed as corollary 4 in~\citet{germain2016pac} which is shown in equation~\ref{eq:PAC_Bayes_bound}. We back up our methodology with some experimental results in section~\ref{sec:experiments_n_results}.
\begin{equation}\label{eq:PAC_Bayes_bound}
    l_{\mathcal{X}}(p(\boldsymbol{w} \mid \mathcal{D})) \leq  \frac{1}{N}\left[l_{\mathcal{D}}(p(\boldsymbol{w} \mid \mathcal{D})) + \mathrm{KL}\left[p(\boldsymbol{w} \mid \mathcal{D})||p(\boldsymbol{w})\right] + \log \frac{1}{\delta} \right] + \frac{s^2}{2}.
\end{equation}
Here, $s^2$ is the \emph{variance factor} of \emph{nll} when treated as a \emph{subgaussian} distribution. A random variable is called subgaussian with variance factor $s^2$ when its tail is dominated by the tail of another gaussian with variance $s^2$~\cite{rivasplata2012subgaussian}. Corollary 5, equation 19 and section A.4 in~\citet{germain2016pac} show that the valid values of $s^2$ depends on the loss function, prior, prediction function, and data distribution. Determining their exact values especially when the predictor is a neural network is hard. Hence, in this work, we approximate $s^2$ by setting its value as the variance of our data likelihood~($p(\mathcal{D}\mid \boldsymbol{w})$). For more information on subgaussian distributions, we refer the readers to~\citet{rivasplata2012subgaussian}.

Minimizing the bound on the right hand side of equation~\ref{eq:PAC_Bayes_bound} is intractable. The idea here is to find a distribution $q$ in a restricted family of distributions $\mathcal{Q}$. Note that, the bound will always hold irrespective of how bad that approximating distribution $q$ is, but we would want to minimize this bound. The idea here is to use VI to optimize it.

\subsection{ELBO and PAC-Bayes are algorithmically the same}\label{methodology:elbo_and_pac_bayes}
Computing the posterior learnt function distribution $p(\boldsymbol{w} \mid \mathcal{D})$ is usually intractable and hard at best. Hence, we approximate it using the variational distribution $q(\boldsymbol{w} \mid \boldsymbol{\phi})$ which is parameterized by its variational parameters $\boldsymbol{\phi}$. On replacing $l_{\mathcal{D}}(p(\boldsymbol{w} \mid \mathcal{D}))$ with our PAC-Bayes risk metric $- \log(\mathcal{D}\mid \boldsymbol{w})$ and shuffling a few terms from equation~\ref{eq:PAC_Bayes_bound}, we get equation~\ref{eq:var_PAC_Bayes_bound}.
\begin{equation}\label{eq:var_PAC_Bayes_bound}
    l_{\mathcal{X}}^{nll}(q(\boldsymbol{w} \mid \boldsymbol{\phi})) \leq \frac{1}{N}\left[-\mathbb{E}_{q(\boldsymbol{w}\mid \boldsymbol{\phi})} \left(- \log p(\mathcal{D}\mid \boldsymbol{w})\right) + \mathrm{KL}\left[(q(\boldsymbol{w} \mid \boldsymbol{\phi}))||p(\boldsymbol{w})\right] + \log \frac{1}{\delta}\right] + \frac{s^2}{2}.
\end{equation}
On carefully observing equation~\ref{eq:var_PAC_Bayes_bound} with the equation for \emph{ELBO}~(equation~\ref{eq:VI_optimization}), one can notice that the differences between the formulations are the constants and a negative sign. Hence, \emph{minimizing the negative of ELBO, should minimize the bound}. We corroborate it with our results in section~\ref{results:elbo_n_pac_bayes} where we show that the \emph{negative of ELBO}  strongly positively correlated with the bound in equation~\ref{eq:var_PAC_Bayes_bound}.

\subsection{Obtaining a generalization bound using VI}\label{methodology:bound_with_VI}
Based on the pointed relation between \emph{ELBO} and our PAC-Bayes formulation as shown in the previous subsection~(\ref{methodology:elbo_and_pac_bayes}), we now propose a training methodology for learning controllers on complex, high-dimensional, non-linear tasks, that can yield a performance bound on unseen situations reliably, and generalize better to unseen situations. 
\subsubsection{Training}\label{methodology:training}
PAC-Bayes uses a data-driven probabilistic function. Inferring this can be a computational challenge when facing complex, high-dimensional data. This is critical for applications where one tries to merge it with neural networks to leverage its universal learnability.

We use a technique for training motivated from ~\citet{blundell2015weight} to evaluate and update \emph{ELBO}. Approximating \emph{ELBO} instead of evaluating it in a closed form allows for a wide variety of prior and posterior distributions. This whole set-up is made amenable for back-propagation using Gaussian re-parameterization~\cite{kingma2015variational}. The core idea behind the re-parametrization trick is to make all randomness an input to the model, making the network deterministic for differentiation. The approximating (variational) distribution $q(\boldsymbol{w}\mid \boldsymbol{\phi})$ over the learnt function is given by a Gaussian with diagonal co-variance, $q(\boldsymbol{w}\mid \boldsymbol{\phi})= \mathcal{N}(\boldsymbol{\mu}, \boldsymbol{\Sigma})$. This Gaussian is parameterized with $\boldsymbol{\phi} = [\boldsymbol{\mu}, \boldsymbol{\rho}]$, where  $\boldsymbol{\Sigma}_{ii}=\log(1+\exp(\boldsymbol{\rho}_i))$. Therefore,  $q(\boldsymbol{w}\mid \boldsymbol{\phi}) = \mathcal{N}(\boldsymbol{w}|\boldsymbol{\mu}, \boldsymbol{\sigma}) = \prod_{w_j\in \boldsymbol{w}} \mathcal{N}(w_j|\mu_j, \sigma_j)$. $\boldsymbol{\Sigma}_{ii}$ is computed in this way to make sure that it is always positive. Fitting the model by VI is done by minimizing the \emph{ELBO} which also minimizes the $\mathrm{KL}$ divergence between the true and approximate posterior~(see equation~\ref{eq:VI_optimization}). It also has an information theoretic justification by means of a bits-back argument~\cite{Hinton:1993:KNN:168304.168306}. Hence, our effective training cost function becomes as shown in equation~\ref{eq:methodology_training}.
\begin{equation}\label{eq:methodology_training}
    \mathcal{F}(\mathcal{D}) \approx \frac{1}{M}\sum_{i=1}^{M} \left(\log q(\boldsymbol{w}^{(i)}\mid \boldsymbol{\phi}) - \log p(\boldsymbol{w}^{(i)}) - \log p(\mathcal{D} \mid \boldsymbol{w}^{(i)}) \right), 
\end{equation}
where, $M$ is the number of Monte-Carlo samples, $\boldsymbol{w}^{(i)} = \mu + (\log(1 + \exp(\rho)) \circ \tau^{(i)})$ and $\tau^{(i)} \sim \mathcal{N}(0, I)$. $\circ$ is an element-wise product. Note that, $\tau^{(i)}$ is implemented as an external input to the learning architecture which is the core trick involved in gaussian reparameterization. This is what makes the whole architecture differentiable with respect to the unbiased Monte-Carlo estimates. Also note that, higher the $M$, higher would be the training stability at the cost of higher training times. We set prior as a zero-mean Gaussian with diagonal unit covariance, i.e. $p(\boldsymbol{w}) = \mathcal{N}(\boldsymbol{0}, \boldsymbol{\mathrm{I}})$. We further use its form that is amenable for training with mini-batches as shown in equation~\ref{eq:methodology_training_minibatches}.
\begin{equation}\label{eq:methodology_training_minibatches}
    \mathcal{F}(\mathcal{D}_{j}) \approx \frac{1}{M}\sum_{i=1}^{M} \left(\theta_{j}(\log q(\boldsymbol{w}^{(i)}\mid \boldsymbol{\phi}) - \log p(\boldsymbol{w}^{(i)})) - \log p(\mathcal{D}_{j} \mid \boldsymbol{w}^{(i)}) \right), 
\end{equation}
where $\theta_{j}= \frac{2^{B-j}}{2^{B}-1}$, $\mathcal{D}_j$ is the $j^{th}$ minibatch, and $B$ is the number of minibatches. We would want to highlight the fact here that the term $\log q(\boldsymbol{w}^{(i)}\mid \boldsymbol{\phi}) - \log p(\boldsymbol{w}^{(i)})$ in equation~\ref{eq:methodology_training} plays an equivalent role of the complexity term $\mathcal{C}$ as described in equation~\ref{eq:PAC_Bayes_definition}. Hence, this brings a regularizing effect to the learner. Note that techniques like this also fall under the umbrella of doubly-stochastic estimation~\citep{titsias2014doubly} where the stochasticity comes from both the minibatches and the Monte-Carlo approximation of expectation.

\subsubsection{Inference}\label{methodology:inference}
Once the variational distribution $q(\boldsymbol{w}\mid \boldsymbol{\phi^*})$ that minimizes equation~\ref{eq:methodology_training_minibatches} is is learnt by backpropagation, predictions on a new input $\boldsymbol{x_*}$ can be done by using equations~~\ref{eq:methodology_prediction}.
\begin{equation}\label{eq:methodology_prediction}
    p(\boldsymbol{a_*}|\boldsymbol{x_*}) \approx \frac{1}{M}\sum_{i=1}^{M} p(\boldsymbol{a_*}|\boldsymbol{x_*},\pi^{i}), \quad \pi^{(i)} \sim q(\boldsymbol{w}\mid \boldsymbol{\phi^*}).
\end{equation}
The predictive distribution $p(\boldsymbol{a_*}|\boldsymbol{x_*})$ is evaluated as a Normal distribution $\mathcal{N}(\boldsymbol{a_*}\mid \boldsymbol{\mu(x_*)}, \boldsymbol{\sigma^{2}(x_{*})\mathrm{I}})$, where $\boldsymbol{\mu(x_*)}$ and $\boldsymbol{\sigma^{2}(x_{*})}$ are evaluated from $M$ number of multiple evaluations. 

\subsubsection{Estimating the generalization risk bound}\label{methodology:bound}
Minimizing equation~\ref{eq:methodology_training_minibatches} automatically minimizes the bound as has been described in section~\ref{methodology:elbo_and_pac_bayes} and further corroborated with results in section~\ref{results:elbo_n_pac_bayes}. Hence, we evaluate the bound by applying $q^{*}(\pi \mid \mathcal{D})$ using samples on the bound form described in equation~\ref{eq:var_PAC_Bayes_bound} to get the generalization risk bound as shown in equation~\ref{eq:evaluating_bound}.
\begin{equation}\label{eq:evaluating_bound}
    l_{\mathcal{X}}^{nll}(q(\boldsymbol{w}\mid \boldsymbol{\phi^*})) \leq \frac{1}{M}\sum_{i=1}^{M} \left[ \frac{\log q(\boldsymbol{w}^{(i)}\mid \boldsymbol{\phi^*}) - \log p(\boldsymbol{w}^{(i)}) - \log p(\mathcal{D} \mid \boldsymbol{w}^{(i)})}{N} \right] + \frac{1}{N}\log \frac{1}{\delta} + \frac{s^2}{2},
\end{equation}
where $\quad \boldsymbol{w}^{(i)} \sim q(\boldsymbol{w}\mid \boldsymbol{\phi^*})$. Note that the bound for any $j^{th}$ minibatch $\mathcal{D}_j$ can be formulated as shown in equation~\ref{eq:evaluating_bound_minibatches}.
\begin{equation}\label{eq:evaluating_bound_minibatches}
    \begin{split}
        l_{\mathcal{X}}^{nll}(q(\boldsymbol{w} \mid \boldsymbol{\phi^*})) \leq \frac{1}{M}\sum_{i=1}^{M} &\left[\frac{\log q(\boldsymbol{w}^{(i)} \mid \boldsymbol{\phi}^*) - \log p(\boldsymbol{w}^{(i)}) - \log p(\mathcal{D}_{j} \mid \boldsymbol{w}^{(i)})}{\left|\mathcal{D}_j\right|}\right] \\
        &+ \frac{1}{\left|\mathcal{D}_j\right|}\log \frac{1}{\delta} + \frac{s^2}{2},\\
    \end{split}
\end{equation}
where, $\left|\mathcal{D}_j\right|$ is the number of data-points in the minibatch $\mathcal{D}_j$.

\section{Experimental Results}\label{sec:experiments_n_results}
We tested our proposed method on $6$ OpenAI Gym environments~\cite{brockman2016gym}: \emph{Humanoid, HalfCheetah, Swimmer, Ant, Hopper, and Walker2d}. In all environments, the goal is to find locomotion controllers to move forward as fast as possible. An additional constraint on \emph{Humanoid} is not to fall over. For these experiments, we used \emph{proximal learnt function optimization~(PPO)}~\cite{PPO} to obtain demonstrator policies. The demonstrations are fed by a demonstrator $\pi_*$ in the form of a dataset $\mathcal{D}=\{(x_i, a_i)\}_{i=0}^{N}$, where $a_i = \pi_*(x_i) \in \mathcal{A}$. We use $10$ episodes of demonstrations to get the results that we show in the following sections from all the experiments that we do. We use a neural network with $3$ hidden layers of sizes $90$, $30$, and $10$ whose weights are parameterized as a Gaussian distribution and optimized with an \emph{Adam}~\citep{kingma2014adam} optimizer on a loss function given by equation~\ref{eq:methodology_training_minibatches} using a learning rate of $0.001$. The bounds are obtained using the optimized variational parameters $\boldsymbol{\phi^*}$ using either equations~\ref{eq:evaluating_bound} or~\ref{eq:evaluating_bound_minibatches}. The probability for the bounds to hold, $\delta$ is set as $0.1$ for all the experiments. This means all our results hold with $0.99$ probability. Note that this can be selected as arbitrarily large. All training is done for $5000$ epochs. The data is fed as $20$ minibatches to the learner. The action taken by controller on a state $\boldsymbol{x_*}$ is $\boldsymbol{\mu(x_*)}$ using equation~\ref{eq:methodology_prediction}. We set $s^2$ and the variance of $p(\mathcal{D} \mid \boldsymbol{w})$ as an hyperparameter that we call as $\beta$ of value $100$.

\begin{wrapfigure}{r}{0.5\textwidth}
  \begin{center}
    \includegraphics[width=0.4\columnwidth]{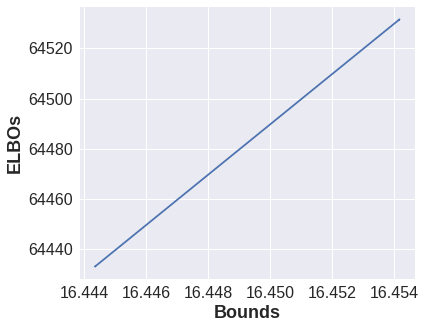}
  \end{center}
  \caption{The numerical values of \emph{ELBO}~(equation~\ref{eq:evaluating_bound}) and our bound~(equation~\ref{eq:var_PAC_Bayes_bound}) increase or decrease with the same rate on the \emph{Swimmer} environment. This follows the fact that \emph{ELBO} and the PAC Bayes formulation with \emph{nll} as the risk metric are algorithmically the same. Additionally, its \emph{R-value} is always $1$ and \emph{P-value} is always $0$. This behavior holds true for all the $6$ \emph{MuJoCo} experiments that we tested our method on.}
    \label{fig:ELBO_bound_correlation}
\end{wrapfigure}

We test our methodology under diverse environments with diverse settings. We first demonstrate the robustness of our methodology by empirically corroborating that \emph{ELBO} and the PAC-Bayes formulation with \emph{nll} as the risk metric are same algorithmically and generate bounds hold in all situations. We then demonstrate the ability of our methodology to generalize to novel situations that are hard to generalize to in general. Finally, we do some sensitivity analysis of methodology in terms of the degree of the fit of the controller to the data. The code for all the experiments is available at \url{https://bit.ly/2qv0JjA}.

\subsection{ELBO and PAC-Bayes are algorithmically the same}\label{results:elbo_n_pac_bayes}
We explained in section~\ref{sec:methodology} that our training objective \emph{ELBO}~(equation~\ref{eq:evaluating_bound}) and our bound equation~(equation~\ref{eq:var_PAC_Bayes_bound}) when used with the \emph{nll} as the risk metric are algorithmically the same. Our experiments in the $6$ \emph{MuJoCo} environments mentioned above prove empirically that it is true. We show one example of all such results in figure~\ref{fig:ELBO_bound_correlation} where we gather the numerical values of \emph{ELBO} and bound using equations~\ref{eq:methodology_training_minibatches} and~\ref{eq:evaluating_bound_minibatches} during the training on the \emph{Swimmer} environment. Additionally, we computed the \emph{pearson correlation coefficient} or the \emph{R-value} and its corresponding \emph{P-values} for the gathered numerical values of \emph{ELBO} and the \emph{bound} in all the environments. \emph{R-value} always came as $1$ and \emph{P-values} as $0$. For the sake of completeness, we show results from all other setting in the supplementary material.

\subsection{Bounds hold always}\label{results:reliable_bounds}
We test the reliability of the bounds by using the demonstrator actions to evaluate the \emph{nll} of the model predictions taken as actions during the validation under diverse settings in all the $6$ \emph{MuJoCo} environments. The bounds are found to hold correctly in all the situations. We show some of them in table~\ref{table:reliable_bounds}. 
 Note that, the value of $s^2$ depends on many factors such as the assumptions on the additive Gaussian noise, iid data etc which we have not considered in our work. Neglecting these terms make the bound tighter than the true bound which should effectively hold with lower than $1-\delta$ probability. Although the generated bounds are found to hold correctly always in our experiments, we suggest incorporating them in practice using some heuristics. However, formulating a relationship between $s^2$ with such assumptions is hard, especially in the context of neural network based controllers that yield an arbitrary distribution.
\begin{table}[H]
\centering
\begin{tabular}{c|c|c|c|c|c|c}
    Metric & Ant & HalfCheetah & Swimmer & Humanoid & Hopper & Walker2d \\
    \hline \hline
    $l_{\mathcal{X}}^{nll}$ & $25.99$ & $19.35$ & $6.46$ & $55.02$ & $9.71$ & $19.39$ \\
    \hline
    Bound & $78.79$ & $70.51$ & $57.54$ & $111.91$ & $61.05$ & $70.7$ \\
    \hline
\end{tabular}
\caption{Our bound estimates hold in all the $6$ MuJoCo environments that we tested it on. \label{table:reliable_bounds}\emph{nll} for validation were obtained using the actions taken by the learner concerning the ones that the demonstrator would have taken. Bounds were obtained using equation~\ref{eq:var_PAC_Bayes_bound} after training using the trained parameters $\boldsymbol{\phi}^*$. More results under diverse settings that further validates the correctness of our bounds can be found in the supplementary file.}
\end{table}

\subsection{Better generalization}\label{results:improved_generalization}
The complexity-term of the \emph{ELBO} which is the cost function that we optimize introduces a regularizing effect to the learning. This can be crucial in situations where an overfitted learnt function might do badly when the situation in hand is different enough from the situations seen during training. Hence, by this complexity-term, our methodology has the potential to generalize better to novel situations. To validate this, we create a different task within the \emph{Swimmer} and \emph{HalfCheetah} environments by changing either or both the lengths and masses of various body parts of the model. These changes are deliberately made in a way that overfitting to the demonstrations from the original task would give poor performance.
 Table~\ref{table:improved_generalization} shows our method achieves more reward in situations which is different enough for the model not to generalize. Although the overall performance decreases, our methodology always performs better than a model that just learns to do well on the given demonstrations. 

\begin{table}[H]
\centering
\begin{tabular}{c|c|c|c}
    Environment & Metric & Original & Other Task \\
    \hline \hline
    \multirow{2}{4em}{Swimmer} & Our technique & $322$ & $133$ \\
    & DNN & $327$ & $62$ \\
    \hline
    \multirow{2}{4em}{HalfCheetah} & Our technique & $5905$ & $2616$ \\
    & DNN & $5833$ & $-18$ \\
    \hline
\end{tabular}
\caption{Our method generalizes better than a \emph{deep neural network}~(DNN) to a novel task that hard to generalize in general. DNN is trained using the mean-squared loss as it loss function and no regularization. \label{table:improved_generalization} 
 Each number under the last two columns are the episodic rewards averaged over $10$ rollouts. The average episodic rewards in the demonstrations fed to the learners are $327$ and $5412$ for Swimmer and HalfCheetah respectively. There is a drop in performance, but our method always does better than a DNN that just overfitted to the original task.}
\end{table}

\subsection{Sensitivity to the degree of dominance of likelihood on the complexity}\label{results:where_it_does_not_work}
The performance of the learner depends on the statistical modeling assumptions such as the choice of the likelihood, the choice of the prior and hyperparameters. The doubly-stochastic estimation from the minibatches and the Monte-Carlo approximation of expectation do not help either. Hence, we do a sensitivity analysis on the learnability of the model based on the degree to which the \emph{nll} dominates the complexity term in the cost function that we optimize~(equation~\ref{eq:methodology_training}). We create three different configurations: C1, C2, and C3. C1 is dominated by the complexity term, C2 is right in between, and C3 is dominated by \emph{nll}. Another way of looking into it is C3 is the least regularized setting of all the three combinations. Table~\ref{table:emphasis_sensitivity} shows that the controller have hard time learning from $\mathcal{D}$ if the \emph{nll} does not dominate our cost function. Another observation here is that sometimes C2 yields better performance than C3~(\emph{Ant}). This indicates that sometimes a regularized learnt function is better than overfitting even on the task on which it has been trained.
\begin{table}[H]
\centering
\begin{tabular}{c|c|c|c}
    Metric & C1 & C2 & C3\\
    \hline \hline
    Swimmer & $20$ & $333$ & $324$ \\
    \hline
    HalfCheetah & $31$ & $4607$ & $5554$ \\
    \hline
    Ant & $777$ & $6389$ & $5976$ \\
    \hline
    Humanoid & $167$ & $926$ & $1938$ \\
    \hline
    Hopper & $63$ & $3485$ & $3438$ \\
    \hline
    Walker2d & $-29$ & $6057$ & $6054$ \\
    \hline
\end{tabular}
\caption{Results showing the learnability of the model and correlation between \emph{ELBO} and PAC-Bayes bound on increasing dominance of the likelihood term on the complexity term from C1 to C3. \label{table:emphasis_sensitivity} It shows that in order to learn from $\mathcal{D}$, the learner needs a likelihood distribution that generates more dominance over the complexity term. Each row is the averaged episodic rewards over $10$ rollouts after training.}
\end{table}
Note that the generalization risks obtained under all settings always hold, irrespective of how good or bad the rewards earned by the learner are. 


\section{Conclusion and Future work}\label{sec:conclusion_n_future_work}
In this work, we build a technique for simultaneously learning a model and a generalization bound from data without making any explicit assumptions about the environment. This is especially important in domains where the datasets are small, but the environments are complex and high-dimensional. We build this technique on top of a relationship we pointed out between \emph{ELBO} and a PAC-Bayes formulation with \emph{nll} as the risk metric. Through our experiments on multiple locomotion tasks on \emph{MuJoCo}, we show the validity of our theory and its robustness to simultaneously learn a learnt function and estimate the generalization risk across diverse settings. We also do a sensitivity analysis of in terms of its learning quality depending on the relative dominance of the likelihood term over the complexity term in the cost function we use for learning.

One legitimate criticism of modelling of $q(\boldsymbol{w}|\boldsymbol{\phi})$ as a Gaussian is it restricts it from learning complex distributions and never converges to $P(\boldsymbol{w}|\mathcal{D})$ even with infinite data. While biases in any model can sometimes be useful, we look forward to integrating \emph{normalizing flows} to approximate distributions of interest with more complex approximations. \emph{Normalizing flows} uses a series of transformations to yield a complex distribution. Having a complex distribution has the potential to yield better policies.

\clearpage

\acknowledgments{We thank NSERC for funding, and Nvidia for their GPUs. We also acknowledge the valuable inputs from Juan Camilog and the anonymous reviewers at CoRL 2019 for their constructive feedback.}



\clearpage
\section*{Supplementary Material}

\section*{Bounds hold always}
We test the validity of the bounds that our method generates on a diverse range of \emph{MuJoCo} locomotion environments under the settings of C1~(table~\ref{table:reliable_bounds_C1}), C2~(table~\ref{table:reliable_bounds_C2}), C3~(table~\ref{table:reliable_bounds_C3}) as explained under section 5.4 in the main paper. For the sake of completeness, the cost function in C1 is dominated by the complexity term, and by the likelihood term in C3. C2 falls right in between.
\begin{table}[H]
\centering
\begin{tabular}{c|c|c|c|c|c|c}
    Metric & Ant & HalfCheetah & Swimmer & Humanoid & Hopper & Walker2d \\
    \hline \hline
    $l_{\mathcal{X}}$ & $25.99$ & $19.35$ & $6.46$ & $55.02$ & $9.71$ & $19.39$ \\
    \hline
    Bound & $78.79$ & $70.51$ & $57.54$ & $111.91$ & $61.05$ & $70.7$ \\
    \hline
\end{tabular}
\caption{Bounds always hold under the settings of C1. \label{table:reliable_bounds_C1} Even though the learnability of the controller is affected under C1 as explain in section 5.4 in the main paper, the bound still holds.}
\end{table}

\begin{table}[H]
\centering
\begin{tabular}{c|c|c|c|c|c|c}
    Metric & Ant & HalfCheetah & Swimmer & Humanoid & Hopper & Walker2d \\
    \hline \hline
    $l_{\mathcal{X}}$ & $25779912$ & $19333044$ & $6443366$ & $65754379$ & $9666841$ & $19333681$ \\
    \hline
    Bound & $26276200$ & $19832960$ & $6943448$ & $55278500$ & $10165692$ & $19833484$ \\
    \hline
\end{tabular}
\caption{Bounds are reliable under the C2 settings. \label{table:reliable_bounds_C2}}
\end{table}

\begin{table}[H]
\centering
\begin{tabular}{c|c|c|c|c|c|c}
    Metric & Ant & HalfCheetah & Swimmer & Humanoid & Hopper & Walker2d \\
    \hline \hline
    $l_{\mathcal{X}}$ & $2.5e12$ & $1.9e12$ & $6.4e12$ & $5.12e12$ & $9.6e12$ & $1.93e12$ \\
    \hline
    Bound & $2.6e12$ & $1.98e12$ & $6.9e12$ & $6.1e13$ & $1.01e13$ & $1.98e12$ \\
    \hline
\end{tabular}
\caption{Bounds are reliable under the C3 settings. \label{table:reliable_bounds_C3}}
\end{table}

\section*{\emph{ELBO} and the PAC Bayes bounds with \emph{nll} as the risk metric are algorithmically the same}
\begin{figure}[!htb]
\minipage{0.32\textwidth}
  \includegraphics[width=\linewidth]{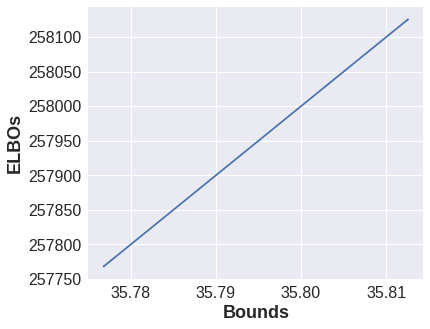}
  \caption{\emph{ELBO} and the bound are positively correlated in \emph{Ant}.}\label{fig:Ant_correlation}
\endminipage\hfill
\minipage{0.32\textwidth}
  \includegraphics[width=\linewidth]{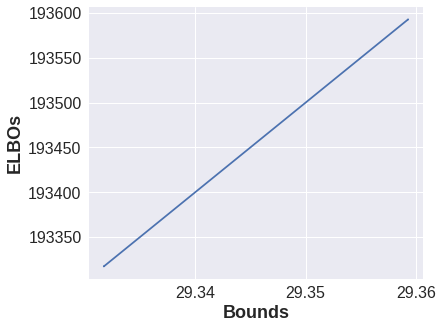}
  \caption{\emph{ELBO} and the bound are positively correlated in \emph{HalfCheetah}.}\label{fig:HalfCheetah_correlation}
\endminipage\hfill
\minipage{0.32\textwidth}%
  \includegraphics[width=\linewidth]{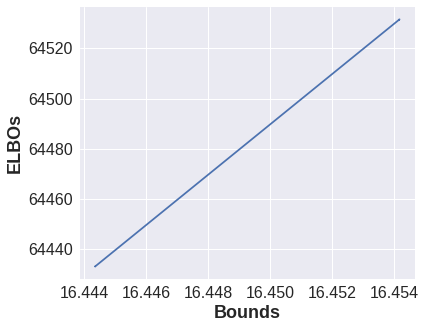}
  \caption{\emph{ELBO} and the bound are positively correlated in \emph{Swimmer}.}\label{fig:Swimmer_correlation}
\endminipage
\end{figure}

\begin{figure}[!htb]
\minipage{0.32\textwidth}
  \includegraphics[width=\linewidth]{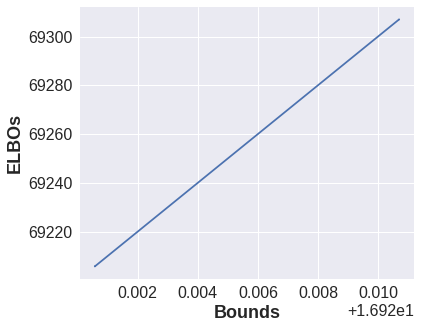}
  \caption{\emph{ELBO} and the bound are positively correlated in \emph{Hopper}.}\label{fig:Hopper_correlation}
\endminipage\hfill
\minipage{0.32\textwidth}
  \includegraphics[width=\linewidth]{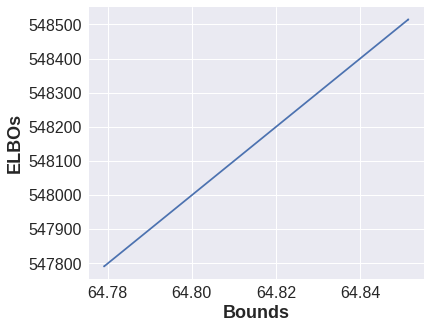}
  \caption{\emph{ELBO} and the bound are positively correlated in \emph{Humanoid}.}\label{fig:Humanoid_correlation}
\endminipage\hfill
\minipage{0.32\textwidth}%
  \includegraphics[width=\linewidth]{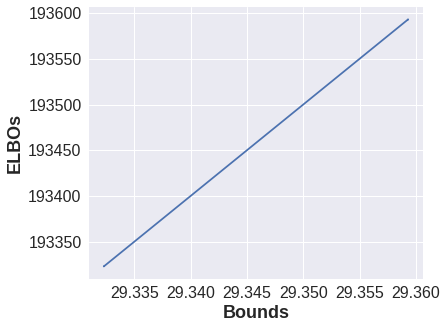}
  \caption{\emph{ELBO} and the bound are positively correlated in \emph{Walker2d}.}\label{fig:Walker2d_correlation}
\endminipage
\end{figure}

We plot the relation between the numerical values of the \emph{ELBO} and the PAC Bayes bound that we obtained during the training on \emph{Ant}~(figure~\ref{fig:Ant_correlation}), \emph{HalfCheetah}~(figure~\ref{fig:HalfCheetah_correlation}), \emph{Swimmer}~(figure~\ref{fig:Swimmer_correlation}), \emph{Hopper}~(figure~\ref{fig:Hopper_correlation}), \emph{Humanoid}~(figure~\ref{fig:Humanoid_correlation}), \emph{Walker2d}~(figure~\ref{fig:Walker2d_correlation}). The experiments are done under C3 settings as described above.

\section*{Manually created MuJoCo task information}
\begin{figure}
    \centering
    \includegraphics[width=0.7\columnwidth]{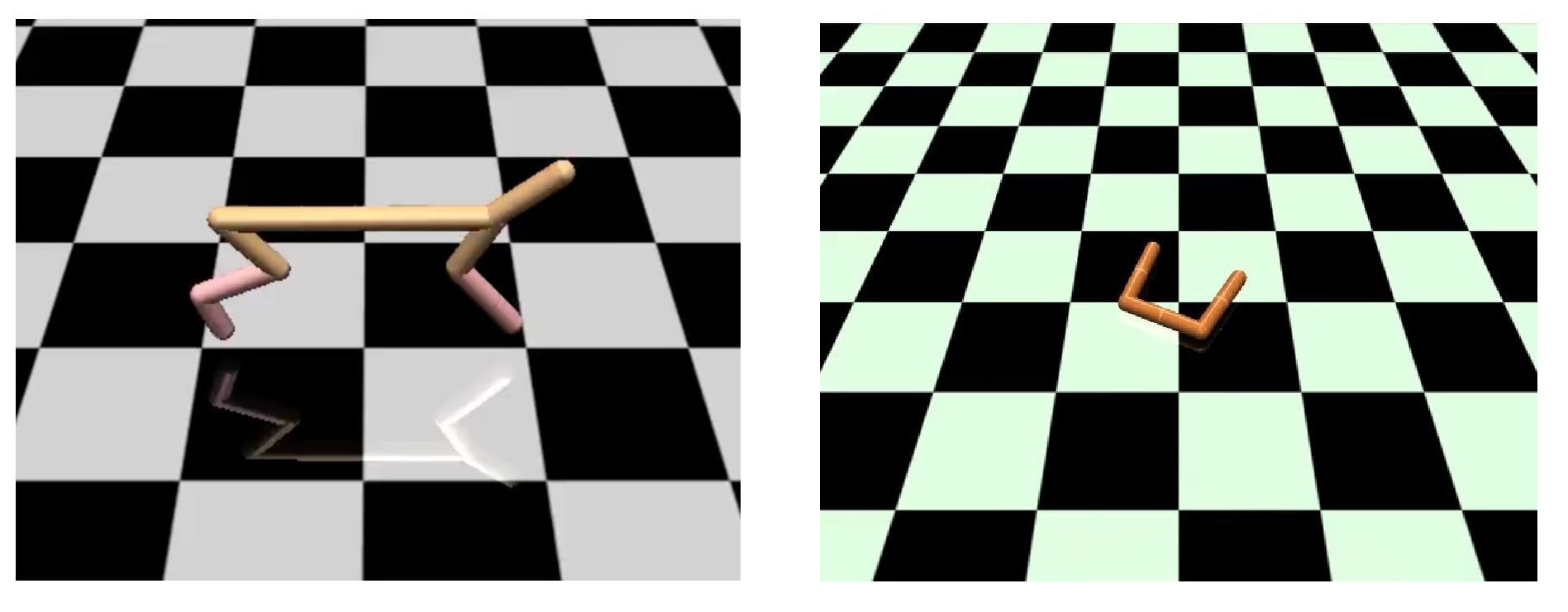}
    \caption{Illustration of the HalfCheetah(left) and Swimmer(right) MuJoCo experiments that we employed in our work.}
    \label{fig:mujoco_tasks}
\end{figure}

We tested our method on two modified OpenAI Gym environments~(\cite{brockman2016gym}) by creating a task on both HalfCheetah and Swimmer. In both environments, the goal is to find locomotion controllers to more forward as fast as possible. To simulate different tasks, we changed either or both the lengths and masses of various body parts like torso, middle, and back. The MuJoCo version used is \emph{mjpro131} with the domains identified as \emph{HalfCheetah-v1} and \emph{Swimmer-v1}. The exact configuration details are specified in tables \ref{table:halfcheetah_task_specifications} and \ref{table:swimmer_task_specifications}. The motive behind setting this experiments is to show the potential of our approach to generalize better to complex and high dimensional tasks than a deterministic neural network with mean-squared loss as the loss function and no regularization.

\begin{table}[H]
\centering
    \begin{tabular}{ c|c|c|c }
    \textbf{Task Identity} & \textbf{Mass dimension(s)} & \textbf{Original Mass} & \textbf{Changed Mass}\\
     \hline \hline
        1       &$(2,3)$   &$(1.53, 1.58)$   &$(0.7, 0.4)$ \\
     \hline
    \end{tabular}
    \caption{Masses of body segments of different tasks of HalfCheetah domain}
    \label{table:halfcheetah_task_specifications}
\end{table}

\begin{table}[H]
\centering
    \begin{tabular}{ c|c|c|c }
    \textbf{Task Identity} & \textbf{Length dimension} & \textbf{Mass dimension} & \textbf{Changed Value}\\
     \hline \hline
        1       &$0$    &$2$       &$(4, 28)$\\
     \hline
    \end{tabular}
    \caption{Masses of body segments of different contexts of Swimmer domain. Note that the default body lengths of the Swimmer is a $3$-dimensional vector of value $(1, 1, 1)$ and default body mass is a $4$-dimensional vector of value $(0, 34.558, 34.558, 34.558)$.}
    \label{table:swimmer_task_specifications}
\end{table}

\end{document}